\documentclass[conference]{IEEEtran}
\usepackage[numbers,sort&compress]{natbib}

\ifCLASSINFOpdf
 \usepackage[pdftex]{graphicx}
\else
\fi

\usepackage{amsmath}

\usepackage{amssymb}

\usepackage{multirow}

\usepackage{algorithm}
\usepackage{algpseudocode}

\usepackage{subfigure}

\begin{document}
	
	\title{Relational Autoencoder for Feature Extraction}
	
	\author{\IEEEauthorblockN{
			Qinxue Meng\IEEEauthorrefmark{1},
			Daniel Catchpoole\IEEEauthorrefmark{2},
			David Skillicorn\IEEEauthorrefmark{3},
			and Paul J. Kennedy\IEEEauthorrefmark{1}}
		\IEEEauthorblockA{\IEEEauthorrefmark{1}
			Centre for Artificial Intelligence, Faculty of Engineering and Information Technology, \\ University of Technology Sydney, Sydney, Australia \\
		}
		\IEEEauthorblockA{\IEEEauthorrefmark{2}
			Children's Cancer Research Unit, The Children's Hospital at Westmead, Sydney, Australia\\}
		\IEEEauthorblockA{\IEEEauthorrefmark{3}
			School of Computing, Queen's University at Kingston, Ontario, Canada\\}
		\IEEEauthorblockA{\IEEEauthorrefmark{0} Email: Qinxue.Meng@uts.edu.au, Daniel.catchpoole@health.nsw.gov.au, skill@cs.queensu.ca, Paul.Kennedy@uts.edu.au}
	}
	
	\maketitle
	
	\begin{abstract}
		Feature extraction becomes increasingly important as data grows high dimensional. Autoencoder as a neural network based feature extraction method achieves great success in generating abstract features of high dimensional data. However, it fails to consider the relationships of data samples which may affect experimental results of using original and new features. In this paper, we propose a Relation Autoencoder model considering both data features and their relationships. We also extend it to work with other major autoencoder models including Sparse Autoencoder, Denoising Autoencoder and Variational Autoencoder. The proposed relational autoencoder models are evaluated on a set of benchmark datasets and the experimental results show that considering data relationships can generate more robust features which achieve lower construction loss and then lower error rate in further classification compared to the other variants of autoencoders.
	\end{abstract}
	
	
	% \fi
	\IEEEpeerreviewmaketitle
	
	\section{Introduction}
	As data becomes increasingly high-dimensional such as genomic information, images, videos and text, reducing dimensionality to generate a high-level representation is considered not only as an important but also necessary data preprocessing step. This is because although machine learning models should, theoretically, be able to work on any number of features, high-dimensional datasets always bring about a series of problems including over-fitting, high-computational complexity and overly complicated models, which gives rise to a well known issue - curse of dimensionality~\cite{indyk1998approximate}. Another reason for reducing dimensionality is that high-level representations can help people better understand the intrinsic structure of data.
	
	Various methods of dimensionality reduction~\cite{wang2015survey, cunningham2015linear, akkarapatty2016dimensionality,WU:ICDM,WU:IJCNN2} have been proposed and they can be roughly divided into two categories: feature selection and feature extraction. The major difference between them lies in using part or all of input features. For example, considering a given dataset~$X$ with a feature set $F$, feature selection finds a subset of features~$D_{s}$ from all features~$F$~(~$D_{s} \subset F$) and the number of selected features is smaller than the original~(~$|D_{s}| \ll |F|$) while feature extraction generates a new set of features~$D_{e}$ which are combinations of the original ones~$F$. Generally new features are different from original features~(~$D_{e} \nsubseteqq F$) and the number of new features, in most cases, is smaller than original features~(~$|D_{e}| \ll |F|$). Although feature selection methods, such as Subset Selection~\cite{john1994irrelevant} and Random Forest~\cite{liaw2002classification}, are effective in filtering out unnecessary features, if all input features contribute to final results to some extent, feature selection is more likely to give rise to information loss than feature extraction because it fails to use all of them. Therefore, the research focus of dimensionality reduction has been moving towards feature extraction, though not totally.
	
	The initial feature extraction methods~\cite{khalid2014survey} are proposed based on projection, mapping input features in the original high-dimensional space to a new low-dimensional space by minimizing information loss. The most famous projection methods are Principal Component Analysis~(PCA)~\cite{demvsar2013principal} and Linear Discriminant Analysis~(LDA)~\cite{sharma2015linear}. The former one is an unsupervised method, projecting original data into its principal directions by maximizing variance. The latter one is a supervised method to find a linear subspace by optimizing discriminating data from classes. The major drawback of these methods is that they do projection linearly. Subsequent studies~\cite{scholkopf1997kernel, lee2004nonlinear, honeine2012online, lebart2013correspondence, lopez2014randomized} overcome this issue by employing non-linear kernel functions.
	
	However, projecting from a high-dimensional space to a low-dimensional one is hard to maintain relationships among data samples which gives rise to change the relationship among data samples so as to generate different results of using original and new features. Therefore, recently, exploiting data relationships has been a major focus of dimensionality reduction research. Multidimensional Scaling~(MDS)~\cite{young2013multidimensional} considers the relationship of data samples by transforming distances into similarities for visualization. ISOMAP~\cite{tenenbaum2000global} learns low-dimensional features by retaining the geodesic distances among data samples. Locally Linear Embedding~(LLE)~\cite{roweis2000nonlinear} preserves data relationships by embedding local neighbourhood when mapping to low-dimensional space. Laplacian Eigenmaps~(LE)~\cite{belkin2003laplacian} minimizes the pairwise distances in the projected space by weighting the corresponding distances in the original space. One issue of these methods is that they generally have a fixed or predefined way of capturing local data relationships in the high-dimensional space which may not be accurate and valid in a low-dimensional space. Another issue is that the major work of these studies maps data from high-dimensional space to low-dimensional space by extracting features once instead of stacking them to gradually generate deeper levels of representation.
	
	%Later, LEE and LE are extended to Neighborhood Preserving Embedding~(NPE)~\cite{1544858} and Locality Preserving Projection~(LPP)~\cite{} to address out-of-sample problem respectively.
	
	Autoencoders~\cite{6302929} use artificial neural networks to reduce dimensionality by minimizing reconstruction loss. Thus, it is easy to stack by adding hidden layers. Because of this, autoencoders and extensions such as Sparse Autoencoders~\cite{deng2013sparse}, Denoising Autoencoders~\cite{vincent2008extracting}, Contractive Autoencoders~\cite{rifai2011contractive} and Variational Autoencoders~\cite{goodfellow2014generative}, demonstrate a promising ability to extract meaningful features, especially in image processing and natural language processing. Yet, these methods only consider data reconstruction and ignore to explicitly model its relationship. A recent study~\cite{wang2014generalized} proposes a Generalized Autoencoder~(GAE) which focuses on modifying autoencoder to consider data relationship by adding a weighted relational function. Specifically, it minimizes the weighted distances between reconstructed instances and the original ones. Although subsequent applications~\cite{camlica2015autoencoding, stober2015deep, gao2015learning, Wu:2014, wang2015dimensionality, meng2016research} confirm that considering data relationship can improve the performance of autoencoder in dimensionality reduction, the Generalized Autoencoder model has some drawbacks. Firstly, determining the weights for each pair-wise distance is very challenging and this practice is risky in converting GAE into a supervised model when some relationshps are over emphasized and others are neglected. Meanwhile focusing on minimizing the loss of data relationships but ignoring the loss of data itself is very likely to lose information. For example, some data attributes contribute more to data relationships than the other. Focusing on reconstructing data relationships may emphasize part of attributes and ignores the others.
	
	To deal with above issues, we proposed a Relation Autoencoder~(RAE) for dimensionality reduction and the contributions are summarized as
	\begin{enumerate}
		\item RAE considers both data features and their relationships by minimizing the loss of them.
		\item To alleviate the increased computational complex of considering data relationships, the proposed Relational Autoencoder model can filter out weak and trivial relationships by adding activation functions.
		\item The relationship model has been extended to work with other baseline autoencoder models including Sparse Autoencoder~(SAE), Denoising Autoencoder~(DAE) and Variational Autoencoder~(VAE).
	\end{enumerate}
	In this paper, we comprehensively evaluate the proposed Relational Autoencoder on a set of benchmark datasets. Specifically, in the experiment, we firstly compare the performance of our proposed model with another recently published relationship-based autoencoder model~(GAE). Then we compare the performance of these baseline autoencoder models with their extended versions.
	
	The rest of the paper starts from reviewing related work of autoencoders in Section~\ref{sec2} followed by problem definition and basic autoencoders in Section~\ref{sec3}. Section~\ref{sec4} presents the proposed Relational Autoencoder model and its extensions of the other baseline autoencoder models. The experimental datasets and results are covered in Section~\ref{sec5} and Section~\ref{sec6} respectively. Finally, we discuss the proposed method and conclude the paper in Section~\ref{sec7}.
	
	\section{Literature review} \label{sec2}
	Autoencoder was initially introduced in the later 1980s~\cite{rumerhart1986learning} as a linear feature extraction method. Unlike latent space approaches which map data into a high dimensional space, autoencoder aims to learn a simpler representation of data by mapping the original data into a low-dimensional space. The main principle of autoencoder follows from the name. ``Auto" presents that this method is unsupervised and ``encoder" means it learns another representation of data. Specifically, autoencoder learns a encoded representation by minimizing the loss between the original data and the data decoded from this representation. In 1989, Baldi and Hornik~\cite{baldi1989neural} investigated the relationship between a one-layer autoencoder and principal component analysis~(PCA). They found that the new compressed features learnt by linear autoencoder is similar to principal components. However, the computational complexity of training an autoencoder is much higher than PCA because the major computational complexity of PCA is based on matrix decomposition. This limits the application of linear autoencoders.
	
	Later, with the involvement of non-linear activation functions, autoencoder becomes non-linear and is capable of learning more useful features~\cite{japkowicz2000nonlinear} than linear feature extraction methods. Non-linear autoencoders are not advantaged than the other non-linear feature extraction methods as it takes long time to train them.
	
	The recent revival of interest in autoencoders is due to the success of effectively training deep architectures because traditional gradient-based optimization strategies are not effective when hidden layers are stacked multiple times with non-linearities. In 2006, Hinton~\cite{hinton2006reducing} empirically trained a deep network architecture by sequentially and greedily optimizing weights of each layer in a Restricted Boltzmann Machine~(RBM) model to achieve global gradient-based optimization. Bengio~\cite{larochelle2007empirical} achieved success in training a stacked autoencoder on the MNIST dataset with a similar greedy layerwise approach. This training approach overcomes the problematic non-convex optimization which prevents deep network structure. Subsequent studies show that stacked autoencoder model can learn meaningful, abstract features and thus achieve better classification results in high-dimensional data, such as images and texts~\cite{jarrett2009best, vincent2009international, ng2016dual, shin2013stacked}. The training efficiency of stacked autoencoder is further improved by changing layer weight initialization~\cite{erhan2010does}.
	
	As the training efficiency improves, autoencoder becomes increasingly popular. Soon it was found that as layers are stacked, the weights of deeper layers increase sharply because of matrix multiplication and then the importances of these weights become larger than the initial input features. This overfitting issue gives rise to the fact that representations of deep layers are more likely dependent on the network structure instead of the initial input features. Poultney~et~al.~\cite{poultney2006efficient} presented an idea to increase the sparsity of network structure so as to limit the increasing of weights. Goodfellow et al.~\cite{goodfellow2009measuring} added a regularization term in the loss function of autoencoder to impose a penalty on large weights. Vincent~et~al.~\cite{vincent2010stacked} proposed a Denoising Autoencoder~(DAE) to solve this issue by adding noises in the input. The proposed DAE model aims not to reconstruct the original input but to reconstruct a corrupted input which is typically corrupted by adding Gaussian noise. The previous studies have no control of the distribution of hidden-layer representations. So Variational Autoencoder~(VAE)~\cite{goodfellow2014generative} is proposed to generate desired distributions of representations in hidden layers.
	
	Autoencoders are an unsupervised dimension reduction process and the reduced high-level features generally contain the major information of original data. This makes autoencoders not sensitive to slight variations. To make them sensitive to slight variations, Rifai~et~al.~\cite{rifai2011contractive} proposed a Contractive Autoencoder~(CAE). In fact, maintaining mutual relationships among data samples works as an effective way of making autoencoders sensitive to slight variations. Because of this, Wang et al.~\cite{wang2014generalized} proposed a Generalized Autoencoder~(GAE) targeting at reconstructing the data relationships instead of the data features. A series of applications~\cite{camlica2015autoencoding, stober2015deep, gao2015learning, wang2015dimensionality, meng2016research} of GAE confirm that maintaining data relationships can achieve better results but results are highly dependent on how to define and choose distance weights. The practice of having sophisticated, pre-defined distance weights is risky in converting GAE into a supervised model because of assigning high weights to selected relationships is very subjective and may be biased. As a result, we propose a Relation Autoencoder~(RAE) for dimensionality reduction by minimising both the loss of data features and relationships.
	
	\section{Preliminaries} \label{sec3}
	This section begins with a definition of feature extraction followed by a description of basic autoencoder models before presenting our proposed Relational Autoencoder~(RAE) model.
	\begin{figure}[!t]
		\centering
		\includegraphics[width=2.5in]{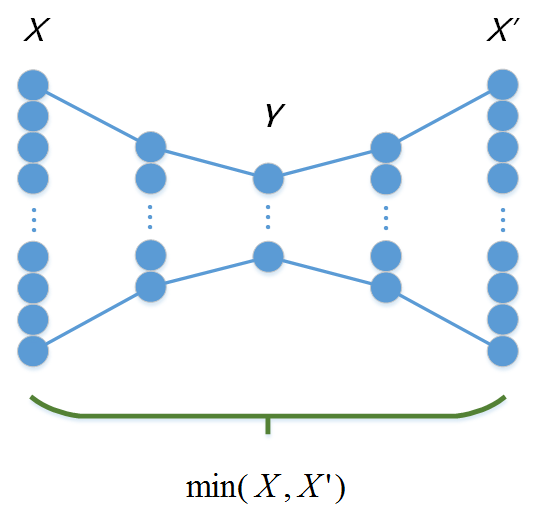}
		\caption{Traditional autoencoder generates new feature set~$Y$ by minimizing the reconstruction loss between $X$ and $X'$.}
		\label{fig_tae}
	\end{figure}
	
	\subsection{Feature Extraction}
	Feature extraction transforms data from the original, high-dimensional space to a relatively low-dimensional space. This transformation can be linear or nonlinear. Specifically, considering a given dataset~$X$ with $n$~samples and $m$~features, the original feature set is denoted as $F_{O}$ and a feature extraction function~$T$ generates new feature set~$F_{N}$ where $|F_{N}|<|F_{O}|$. Generally, the objective functions of feature extraction methods minimize the difference between the original space $F_{O}(X)$ and the new space $F_{N}(X)$ and changing the objective functions can convert feature extraction from unsupervised methods to supervised methods such as Linear Discriminant Analysis~(LDA).
	
	\begin{figure}[!t]
		\centering
		\includegraphics[width=2.5in]{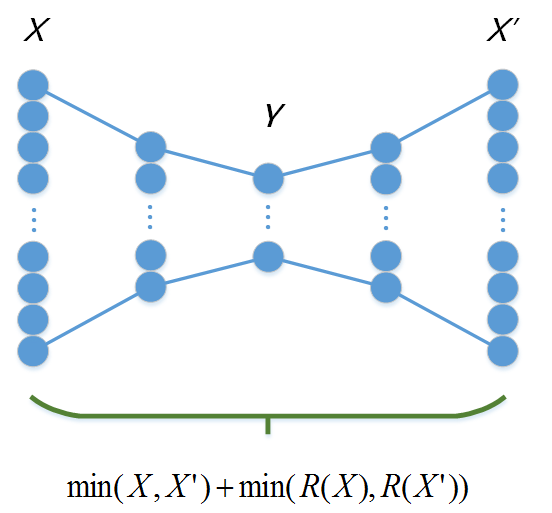}
		\caption{Relational autoencoder generates new feature set~$Y$ by minimizing the loss of reconstructing $X$ and maintaining relationships among $X$.}
		\label{fig_nae}
	\end{figure}
	
	\subsection{Basic Autoencoder}
	Simply, an autoencoder~(AE) is composed of two parts, an encoder and a decoder. Considering a data sample $X$ with~$n$ samples and~$m$ features, the output of encoder~$Y$ represents the reduced representation of~$X$ and the decoder is tuned to reconstruct the original dataset~$X$ from the encoder's representation~$Y$ by minimizing the difference between~$X$ and~$X'$ as illustrated in Fig.~\ref{fig_tae}. Specifically, the encoder is a function~$f$ that maps an input~$X$ to hidden representation~$Y$. The process is formulated as
	\begin{equation}
	Y = f(X) = s_{f}(WX + b_{X})
	\end{equation}
	where $s_{f}$ is a nonlinear activation function and if it is an identity function, the autoencoder will do linear projection. The encoder is parameterized by a weight matrix $W$ and a bias vector~$b~\in~R^{n}$.
	
	The decoder function $g$ maps hidden representation $Y$ back to a reconstruction $X'$:
	\begin{equation}
	X' = g(Y) = s_{g}(W'Y + b_{Y})
	\end{equation}
	where $s_{g}$ is the decoder's activation function, typically either the identity (yielding linear reconstruction) or a sigmoid. The decoder's parameters are a bias vector $b_{y}$ and matrix $W'$. In this paper we only explore the tied weights case where~$W'~=~W^{T}$.
	
	Training an autoencoder involves finding parameters $\theta=~(W, b_{X}, b_{Y})$ that minimize the reconstruction loss on the given dataset $X$ and the objective function is given as
	\begin{equation}
	\Theta = \min_{\theta} L(X, X') = \min_{\theta} L(X, g(f(X)))
	\end{equation}
	For linear reconstruction, the reconstruction loss~($L_{1}$) is generally from the squared error:
	\begin{equation}
	L_{1}(\theta) = \sum_{i=1}^{n} ||x_{i}-x'_{i}||^{2} = \sum_{i=1}^{n} ||x_{i}-g(f(x_{i}))||^{2}
	\end{equation}
	For nonlinear reconstruction, the reconstruction loss~($L_{2}$) is generally from cross-entropy:
	\begin{equation}
	L_{2}(\theta) = -\sum^{n}_{i=1}\left[ x_{i}\log(y_{i}) + (1-x_{i})\log(1-y_{i}) \right]
	\end{equation}
	where $x_{i} \in X$, $x'_{i} \in X'$ and $y_{i} \in Y$.
	
	\subsection{Stacked Autoencoder}
	As a neural network model based feature extraction method, a major advantage of Autoencoder is that it is easy to stack for generating different levels of new features to represent original ones by adding hidden layers. For an $l$-layer stacked autoencoder, the process of encoding is
	\begin{equation}
	Y = f_{l}(\ldots f_{i}( \ldots f_{1}(X)))
	\end{equation}
	where $f_{i}$ is the encoding function of layer $i$. The corresponding decoding function is
	\begin{equation}
	X' = g_{l}(\ldots g_{i}( \ldots g_{1}(Y)))
	\end{equation}
	where $g_{i}$ is the decoding function of layer $i$ and the Stacked Autoencoder can be trained by greedy layerwise feed-forward approach.
	
	\section{Relational Autoencoder~(RAE)}  \label{sec4}
	The traditional autoencoder generates new features by minimizing the reconstruction loss of the data. This motivates us to propose a Relational Autoencoder~(RAE) to minimize the reconstruction loss of both data and their relationships. The objective function of RAE is defined as
	\begin{equation}
	\Theta = (1-\alpha)\min_{\theta} L(X, X') + \alpha\min_{\theta}L(R(X), R(X'))
	\end{equation}
	where $R(X)$ represents the relationship among data samples in $X$ and $R(X')$ represents the relationship among data samples in $X'$ as illustrated in Fig.~\ref{fig_nae}. Parameter $\alpha$ is a scale parameter to control the weights of the data reconstruction loss and the relationship reconstruction loss and $\theta$ is the neural network parameter of the autoencoder.
	
	Data relationship can be modelled in multiple ways. In this paper, we present data relationship by their similarities where $R(X)$ is the multiplication of $X$ and $X^{T}$. Then the objective function is
	\begin{equation}
	\Theta = (1-\alpha)\min_{\theta} L(X, X') + \alpha\min_{\theta}L(XX^{T}, X'X'^{T})
	\end{equation}
	
	In order to improve computational efficiency and filter out unnecessary relationships, we use activation functions to control weights of similarities. In this paper, we use the rectifier function~\cite{lecun2015deep} to achieve this.
	\begin{equation}
	\tau_{t}(r_{ij}) =
	\begin{cases}
	r_{ij}, & \text{if}\ r_{ij} \geq t,  \\
	0,      & \text{otherwise},
	\end{cases}
	\end{equation}
	where $t$ is a threshold to filter out weak and trivial relationships. Then the objective function of RAE is defined as
	\begin{equation}
	\label{fequ}
	\begin{split}
	\Theta &= (1-\alpha)\min_{\theta} L(X, X') \\
	& + \alpha\min_{\theta}L(\tau_{t}(XX^{T}), \tau_{t}(X'X'^{T}))
	\end{split}
	\end{equation}
	In this paper, we choose the loss function~$L$ as squared error.
	
	The pseudo-code of the proposed Relational Autoencoder~(RAE) is described in Algorithm~\ref{alg1} where the input parameters are the input dataset~($X$), parameter of the rectifier function~($t$), regularization weight~($\lambda$), the number of hidden neurons in layers~($N$), activation function~($s_{f}$) and a threshold~($\varepsilon$) to determine whether the loss has converged. Among them, $N$ is a vector and $n_{i}$ is the number of neurons of $i$th~layer where $i$ is from $1$ to $|N|$. The proposed RAE model starts with defining a loss function~($L$). Then it initializes a neural network model and iteratively add layers into the network model based on $N$ and~$s_{f}$. The network is trained by stochastic gradient descent~(SGD) and updates~$\theta$ in the loss function~($L$) until the difference between current-loop loss~(loss\_c) and the previous-loop loss~(loss\_p) is smaller than a predefined threshold~($\varepsilon$) which means it has converged.
	
	\begin{algorithm}[t]
		\caption{Iterative learning procedure of RAE}
		\label{alg1}
		\begin{algorithmic}[1]
			\Function{RAE}{$X, t, \lambda, N_{l}, s_{f}, \varepsilon$}
				\State L = Equ.~\ref{fequ};
				\State loss\_p = 0;
				\State loss\_c = 0;
				\State Model = NN();  \Comment{initialize a neural network model}
				\For{$i=1$ to $|N|$}
					\State Model.addLayer($n_{i}$, $s_{f}$)
				\EndFor
				\While{True}
					\State loss\_c = Model.train(L, SGD);
					\If{(loss\_c - loss\_p $\leqslant$ $\varepsilon$)}
						\State Break;
					\Else
						\State loss\_p = loss\_c;	
					\EndIf
				\EndWhile \\
				\Return Model
			\EndFunction
		\end{algorithmic}
	\end{algorithm}
	
	\subsection{Extension to Sparse Autoencoder~(SAE)}
	The objective function of autoencoder reconstructs the input. During the training process, the weights of hidden neurons are the combination of previous layers and these weights increase as layers get deep. High weights of hidden layers make the generated features more dependent on the network structure rather than the input. In order to avoid this, Sparse Autoencoder~(SAE) imposes weight-decay regularization so as to keep neuron weights small. The objective function of SAE is
	\begin{equation}
	\Theta = \alpha\min_{\theta} L(X, X') + \lambda||W||^{2}
	\end{equation}
	where~$||W||^{2}$ is a weight-decay regularization term to guarantee weight matrix~$W$ having small elements. Parameter~$\lambda$ is a hyper-parameter to control the strength of the regularization. We extend Sparse Autoencoder~(SAE) to a Relational Sparse Autoencoder~(RSAE) model by considering the data relationships. The objective function is defined as
	\begin{equation}
	\begin{split}
	\Theta &= (1-\alpha)\min_{\theta} L(X, X') \\
	& + \alpha\min_{\theta}L(\tau_{t}(XX^{T}), \tau_{t}(X'X'^{T})) \\
	& + \lambda||W||^{2}
	\end{split}
	\end{equation}
	
	\subsection{Extension to Denoising Autoencoder~(DAE)}
	Denoising Autoencoder~(DAE) is proposed to generate better features by imposing an alternative form of regularization. The main principle of DAE is to corrupt a part of the input features of a given dataset $X$ before sending it to an autoencoder model, train the network to reconstruct the destroyed input $\tilde{X}$ and then minimize the loss between the reconstructed $\tilde{X'}$ and original $X$. The objective function of DAE is
	\begin{equation}
	\Theta = \min_{\theta}[L(X, g(f(\tilde{X})))] \hspace{2mm} \text{s.t.} \hspace{2mm}
	\tilde{X} \sim q(\tilde{X}|X)
	\end{equation}
	where $\tilde{X}$ is corrupted $X$ from a stochastic corruption process $q(\tilde{X}|X)$ and the objective function is optimized by stochastic gradient descent. Here we extend the proposed Relation Autoencoder model to a Relational Denoising Autoencoder~(RDAE) model by considering data relationships. The model minimizes the loss between data~$X$ and corrupted data~$\tilde{X}$ and the loss between data relationship~$XX^{T}$ and corrupted data relationship~$\tilde{X}\tilde{X}^{T}$. The objective function is defined as
	\begin{equation}
		\begin{split}
			\Theta &= (1-\alpha)\min_{\theta} L(X, g(f(\tilde{X})) \\
			& + \alpha\min_{\theta}L(\tau_{t}(XX^{T}), \tau_{t}(\tilde{X}\tilde{X}^{T})) \\
			& \hspace{3.2cm} \text{s.~t.} \hspace{2mm}\tilde{X} \sim q(\tilde{X}|X)
		\end{split}
	\end{equation}
	
	In this paper, we consider corruptions as additive isotropic Gaussian noise: $\tilde{X}=X+\varDelta$ where $\tilde{X} \sim N(0, \delta^{2})$ and $\delta$ is the standard deviation of $X$.
	
	\subsection{Extension to Variational Autoencoder~(VAE)}
	Variational Autoencoder~(VAE) is a different to the other types of autoencoder model. It makes a strong assumption concerning the distribution of latent neurons and tries to minimize the difference between a posterior distribution and the distribution of latent neurons with difference measured by the Kullback-Leibler divergence~\cite{kullback1951information}. The objective function of VAE is
	\begin{equation}
	\Theta = \min D_{KL}(q_{\phi}(Y|X)||p_{\theta}(X|Y))
	\end{equation}
	where $q_{\phi}(Y|X)$ is the encoding process to calculate the probability of $Y$ based on the input $X$ while $p_{\theta}(X|Y)$ is the decoding process to reconstruct $X$. Generally $Y$ is a predefined Gaussian distribution, such as $N(0,1)$. Therefore the extended Relational Variational Autoencoder~(RVAE) is defined as
	\begin{equation}
	\begin{split}
	\Theta &= (1-\alpha)\min D_{KL}(q_{\phi}(Y|X)||p_{\theta}(X|Y))\\
	& + \alpha\min D_{KL}(q_{\phi}(Y|XX^{T})||p_{\theta}(XX^{T}|Y))
	\end{split}
	\end{equation}
	
	\section{Datasets}  \label{sec5}
	The datasets used in this paper to evaluate the proposed Relational Autoencoder are two image datasets, MNIST\footnote{http://yann.lecun.com/exdb/mnist/} and CIFAR-10\footnote{https://www.cs.toronto.edu/~kriz/cifar.html}. The MNIST dataset is a well known database of handwritten digits which contains a training set of 60,000 examples and a test set of 10,000 samples. The CIFAR-10 dataset contains 60,000 images which are labelled with 10 classes with each class having 6,000 images. The training set of CIFAR-10 has 50,000 samples while the test set has 10,000 samples.
	
	\section{Experiment} \label{sec6}
	The experiment firstly compares the performance of the proposed Relational Autoencoder~(RAE) model against basic autoencoder~(BAE) and Generative Autoencoder~(GAE) in reconstruction loss and classification accuracy \cite{Wu:IJCNN1,Wu:TNNLS}. Then we compare the performance of the extended Relational Sparse Autoencoder~(RSAE), Relational Denoising Autoencoder~(RDAE) and Relational Variational Autoencoder~(RVAE) with their corresponding original versions to estimate the effects of considering relationship in the depth.
	
	\begin{figure*}[!t]
		\centering
		\subfigure[MNIST]{\includegraphics[width=0.45\textwidth]{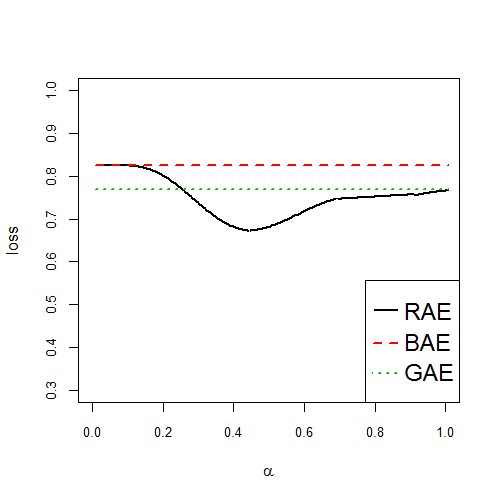}}
		\subfigure[CIFAR-10]{\includegraphics[width=0.45\textwidth]{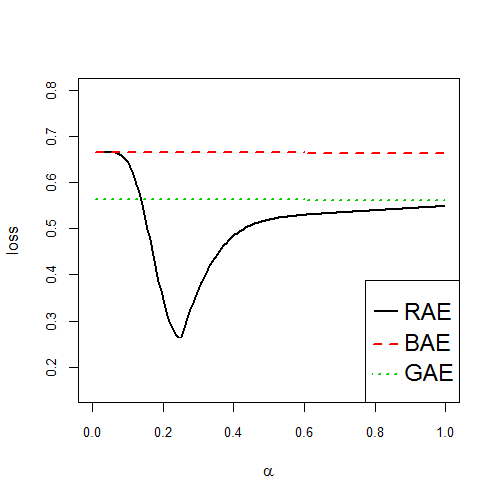}}
		\caption{The comparative results of reconstruction loss on the MNIST and CIFAR-10 dataset among Relational Autoencoder~(RAE), basic autoencoder~(BAE) and Generative Autoencoder~(GAE). It describes how information loss changes as the scale parameter $\alpha$ changes.}
		\label{fig1}
	\end{figure*}
	
	\subsection{Experiment Setting}
	All autoencoder models were tested with the same configuration on the same dataset. Specifically, we use tied weights ~($W' = W^{T}$) in building network structure. The activation function of each layer is sigmoid for both encoder and decoder. The optimization functions of reconstruction loss are listed and described in Section~\ref{sec4} and they are trained by stochastic gradient descent~(SGD) for 400 epochs. The number of neurons of each layer is determined as
	\begin{equation}
		n_{i+1} =
		\begin{cases}
			\log(n_{i}), & \text{if}\ n_{i+1} \geq l_{t},  \\
			 l_{t},          & \text{otherwise},
		\end{cases}
	\end{equation}
	where $n_{i}$ is the number of neurons in layer $i$ and $n_{i+1}$ is the number of neurons in layer $i+1$. As the network structure goes deeper, the number of neurons gradually decreases  to a predefined threshold $l_{t}$. To increase the training efficiency, we use Xavier~\cite{glorot2010understanding} method to initialize layer weights
	\begin{equation}
	w_{ij} = U\left[-\frac{1}{\sqrt{n_{i-1}}}, \frac{1}{\sqrt{n_{i-1}}}\right]
	\end{equation}
	where $w_{ij}$ is weight of $j$th neuron in layer~$i$  and $U[-\frac{1}{\sqrt{n}}, \frac{1}{\sqrt{n}}]$ is the uniform distribution in the interval~($-\frac{1}{\sqrt{n}}$,$\frac{1}{\sqrt{n}}]$) and $n_{i-1}$ is the number of neurons in the previous layer.
	
	All autoencoders are trained based on their loss function respectively. But in order to compare the performance of them in feature extraction, the reconstruction loss is measured by Mean Squared Error~(MSE). Classification is done by softmax regression based on the extracted features from autoencoder models. To estimate the generation of models, we use 10-fold cross validation in both unsupervised feature extraction and classification.

	\subsection{Comparing to BAE and GAE}
	We firstly compare the performance of our proposed Relational Autoencoder~(RAE) with basic autoencoder~(BAE) and Generative Autoencoder~(GAE) in terms of reconstruction on MNIST and CIFAR-10. For GAE, we set similarity to be the weight of each pairwise relationship. In the experiment, we explore different values of scaling parameter~$\alpha$ which determines the weights of reconstructing data and relationship~(Equation~\ref{fequ}). The value of~$\alpha$ ranges from 0 to 1 in step of 0.02. Meanwhile, because BAE and GAE has no such parameter, their reconstruction loss is not changed as~$\alpha$ changes and the results are illustrated in~Fig.~\ref{fig1}.
	
	We observe that generally the reconstruction loss of GAE is less than BAE which confirms that considering data relationship is able to reduce information loss in the process of encoding and decoding. The performance of the proposed RAE autoenocder model changes as the scaling parameter~$\alpha$ changes. Generally, it starts to generate similar results with BAE because $\alpha$ is small focusing on reconstructing data rather than relationships. As~$\alpha$ increases, the performance of GAE continuously decreases until to a tough. Then as~$\alpha$ keeps increasing, the information loss increases as well because the model begins to over-emphasize relationship reconstruction. It is interesting to see that even if the proposed RAE model considers relationship only, the performance of RAE is better than GAE as RAE uses the activation function to filter out weak relationships. Thus the performance of the proposed RAE autoencoder model is determined by scaling parameter~$\alpha$ and the value of~$\alpha$ depends on the dataset.
	
	Another interesting finding is that the proposed RAE model achieves better results in CIFAR-10 than MNIST. This may be because the CIFAR-10 contains more complex images than MNIST and for complex datasets, maintaining data relationship is of much importance. For classification, RAE achieves the lowest error rate~(3.8\%) followed by GAE~(5.7\%) and BAE~(8.9\%).
	
	\begin{table}[t]
		\renewcommand{\arraystretch}{1.5}
		\centering
		\caption{Experimental results of autoencoder models}
		\label{my-label}
		\begin{tabular}{|c|c|c|c|c|}
			\hline
			\multirow{2}{*}{\textbf{Model}}
			& \multicolumn{2}{l|}{\textbf{\hspace*{0.6cm}MNIST}}
			& \multicolumn{2}{l|}{\textbf{\hspace*{0.3cm}CIFAR-10}} \\ \cline{2-5}
			& \textbf{Loss} & \textbf{Error}
			& \textbf{Loss} & \textbf{Error}       \\ \hline \hline
			\textbf{RAE}  & 0.677 & 3.8\% & 0.281 & 12.7\% \\ \hline
			\textbf{BAE}  & 0.813 & 8.9\% & 0.682 & 15.6\% \\ \hline
			\textbf{GAE}  & 0.782 & 5.7\% & 0.574 & 14.9\% \\ \hline \hline
			\textbf{RSAE} & 0.296 & 1.8\% & 0.292 & 13.4\% \\ \hline
			\textbf{SAE}  & 0.312 & 2.2\% & 0.331 & 14.2\% \\ \hline \hline
			\textbf{RDAE} & 0.217 & 1.1\% & 0.216 & 10.5\% \\ \hline
			\textbf{DAE}  & 0.269 & 1.6\% & 0.229 & 11.7\% \\ \hline \hline
			\textbf{RVAE} & 0.183 & 0.9\% & 0.417 & 17.3\% \\ \hline
			\textbf{VAE}  & 0.201 & 1.2\% & 0.552 & 21.2\% \\ \hline \hline
		\end{tabular}
	\end{table}
	
	\subsection{Comparing extend autoencoder variations to original ones}
	In this experiment, we compare the performance of the extended variants of autoencoders to their original versions and detailed results are listed in Table~\ref{my-label}. We observe that considering data relationships contributes to decreasing reconstruction loss suggesting that autoencoders can generate more robust and meaningful features with less information loss and these features are the key to achieve good classification results.
		
	\section{Conclusion} \label{sec7}
	In this paper, we propose a Relation Autoencoder model which can extract high-level features based on both data itself and their relationships. We extend this principle to other major autoencoder models including Sparse Autoencoder, Denoising Autoencoder and Variational Autoencoder so as to enable them consider data relationships as well. The proposed relational autoencoder models are evaluated on MNIST and CIFAR-10 datasets and the experimental results show that considering data relationships can decrease reconstruction loss and therefore generate more robust features. Better features contribute to improved classification results.
	
	
	
	%\bibliographystyle{IEEEtran}
	\bibliographystyle{IEEEtran}
	\bibliography{myReferences}

\end{document}